\documentclass[runningheads]{llncs}
\usepackage{graphicx}

\usepackage{multirow}
\usepackage{booktabs}
\usepackage{amsmath}
\usepackage{subcaption}
\usepackage{hyperref}
\usepackage{orcidlink}

\begin{document}
\title{nnOOD: A Framework for Benchmarking Self-supervised Anomaly Localisation Methods}
%
%

\author{Matthew Baugh\inst{1}\orcidlink{0000-0001-6252-7658} \and
Jeremy Tan\inst{1}\orcidlink{0000-0002-9769-068X} \and Athanasios Vlontzos\inst{1}\orcidlink{0000-0002-7672-2574} \and
Johanna P. M\"uller\inst{2}\orcidlink{0000-0001-8636-7986} \and
Bernhard Kainz\inst{1,2}\orcidlink{0000-0002-7813-5023}}

\authorrunning{M. Baugh et al.}

\institute{Imperial College London, SW7 2AZ, London, UK \\
\email{matthew.baugh17@imperial.ac.uk}
\and Friedrich--Alexander University Erlangen--N\"urnberg, DE}

\maketitle              

\begin{abstract}
The wide variety of in-distribution and out-of-distribution data in medical imaging makes universal anomaly detection a challenging task.
Recently a number of self-supervised methods have been developed that train end-to-end models on healthy data augmented with synthetic anomalies.
However, it is difficult to compare these methods as it is not clear whether gains in performance are from the task itself or the training pipeline around it.
It is also difficult to assess whether a task generalises well for universal anomaly detection, as they are often only tested on a limited range of anomalies.
To assist with this we have developed nnOOD, a framework that adapts nnU-Net to allow for comparison of self-supervised anomaly localisation methods.
By isolating the synthetic, self-supervised task from the rest of the training process we perform a more faithful comparison of the tasks, whilst also making the workflow for evaluating over a given dataset quick and easy.
Using this we have implemented the current state-of-the-art tasks and evaluated them on a challenging X-ray dataset.

\end{abstract}
\section{Introduction}

\parskip5pt

Out-of-distribution detection, \emph{i.e.} learning a normative distribution from a single class and classifying anomalous test samples without supervised training, is a notoriously challenging task. Many methods struggle in scenarios where humans can easily detect an outlier. For example, failures in detecting potholes under different lighting conditions \cite{potholes}, a jet ski in a road scene \cite{chan2021segmentmeifyoucan} or determining whether a picture of a cat is a dog \cite{golan2018deep}.
It becomes even more difficult in medical imaging, where abnormalities frequently go unseen by expert observers, often because of inattentional blindness~\cite{drew2013invisible}.
This issue worsens in high-pressure environments such as emergency and trauma care.
There, the main cause of misdiagnosis is the misinterpretation of radiographs, with miss rates of up to 80\%~\cite{errorsinimaging}.
In these situations an automated tool that acts as a second reader, alerting clinicians to any unusual features, would be useful.

Many publications in the field of medical anomaly detection limit their experiments to datasets with a narrow range of abnormalities \cite{scalespaceAE,vqvaeanomaly,fanogan,tan2020detecting}, raising questions regarding their ability to generalise to anomalies seen in other medical applications or modalities.
Recently there has been a trend of training end-to-end anomaly detection models using synthetic anomalies to alter healthy data.
These methods exhibit good performance on both manufacturing \cite{schluter2021self} and medical tasks \cite{tan2020detecting,PII}, with the majority of the top submissions to the recent MICCAI 2021 Medical Out-of-Distribution challenge (MOOD) \cite{mood_2021_4573948} being trained in this manner.
However, as these tasks are synthetic by nature, it is even more important that they are thoroughly tested, as there is a risk of the task being overly tuned to the target evaluation dataset.

\noindent\textbf{Contribution: } In this paper we present the nnOOD framework.
It builds on top of the nnU-Net framework \cite{Isensee2021}, maintaining the core principle of adapting the architecture to the provided dataset, but gearing it toward self-supervised anomaly localisation.
This makes validating a synthetic anomaly task on a given dataset as simple as organising a dataset for nnU-Net.
We also provide a common interface for self-supervised tasks, simplifying the process of creating and applying a new method.
We hope that making this paradigm of anomaly detection more accessible can encourage more interest in this research area.
Ultimately, we hope that nnOOD provides a standardised framework for comparing these methods fairly.
Another goal is to gain insight into what makes a synthetic task useful and elucidate why a certain task fails or succeeds in different situations.
Source code, including a guide on applying the framework to a new dataset, is available at \href{https://github.com/matt-baugh/nnOOD}{https://github.com/matt-baugh/nnOOD}.

\noindent\textbf{Related Work: } A common way to perform anomaly localisation is by measuring the difference between an image and its reconstruction.
The assumption is that a model trained on normal data will not be able to reproduce anomalous regions, leading them to larger deviations.
This method lends itself most easily to autoencoder-based architectures \cite{scalespaceAE,vqvaeanomaly}, but has also been applied by using generative adversarial networks \cite{han2021madgan,fanogan}.
However, reconstruction loss often fails as models are not always able to reconstruct healthy regions in potentially unhealthy samples \cite{autoencodersForBrainMRI}, and anomalies with extreme textures but a normal intensity distribution are difficult to identify \cite{meissen2022on}.
Sample-level anomaly detection can be done as an auxiliary task using a classification network, although performance can vary greatly between training epochs \cite{10.1007/978-3-030-87735-4_10}.
Methods using normalizing flows \cite{gudovskiy2022cflow,yu2021fastflow} are currently the state-of-the-art benchmarks on the MVTec~\cite{mvtec} dataset.
Unfortunately, these strategies require pre-trained model which is often not possible for medical imaging tasks, and high performance on computer vision tasks does not guarantee similar performance on medical imaging tasks \cite{10.1007/978-3-030-87735-4_12}.

Foreign Patch Interpolation (FPI) \cite{tan2020detecting} was the first method to train end-to-end models for out-of-distribution localisation using synthetic anomalies.
FPI creates subtle anomalies by interpolating a patch of the current sample with a patch extracted from the same location of a different sample.
The model is then trained to predict the pixel-wise interpolation factor, causing the model to learn a score correlated with how anomalous the pixel was.
This had good performance on both brain MRI and abdominal CT data, winning the MICCAI 2020 MOOD challenge \cite{mood_2021_4573948} in both the sample-wise and pixel-wise categories.
Poisson image interpolation (PII) was introduced to mitigate the discontinuities at the edges of FPI's patches \cite{PII}, allowing for a more seamless blend between the patch and its surroundings.
However, there was still limited variation in the synthetic anomalies, due to the interpolated patches being extracted from the same location in the secondary image.
Independently, CutPaste \cite{li2021cutpaste} used a similar sort of augmentation by copying a patch from one place and pasting it at a random different location within the same image.
The patches are sometimes altered through rotation or jitter in the pixel values.
Rather than training end-to-end, \cite{li2021cutpaste} trained a one-class classifier, using a Gaussian density estimator on the output to enable outlier detection, and applying GradCAM \cite{selvaraju2017grad} for anomaly localisation.
Natural synthetic anomalies (NSA) \cite{schluter2021self} combines the aforementioned image augmentations, further increasing the variety of synthetic samples by resizing the patches and randomising the number of patches introduced in an image.
Arguing that in previous methods the difference between the distributions of the blended patches can cause the same label to be applied to vastly different levels of abnormality, they opted to use a scaled logistic function applied to the mean absolute intensity difference across each channel.

The results of the MICCAI 2021 MOOD challenge \cite{mood_2021_4573948} displayed the success of these techniques, with the majority of the presented works being models trained end-to-end with synthetic anomalies.
Despite their achievements, many of those methods faced practical challenges in engineering the synthetic task.
For example, many 3D methods used image resizing which leads to loss of information and obscures small anomalies.
In addition, some methods apply non-overlapping patches to the larger abdominal data, resulting in prediction artefacts around the edges.

We see the structure of nnU-Net \cite{Isensee2021} as a natural solution to these issues.
The nnU-Net framework uses a set of heuristic rules to dynamically adapt a U-Net \cite{unet} to a given biomedical image segmentation dataset.
Combining these architectural decisions with a solid pipeline of adaptive preprocessing, extensive data augmentation, model ensembling and aggregating tiled predictions, nnU-Net consistently performs well across a wide range of tasks.
The ease with which the framework can be applied has made semantic segmentation more accessible, even to those without machine learning expertise.

\section{Method}

\begin{figure}[ht]
    \centering
    \includegraphics[width=\linewidth]{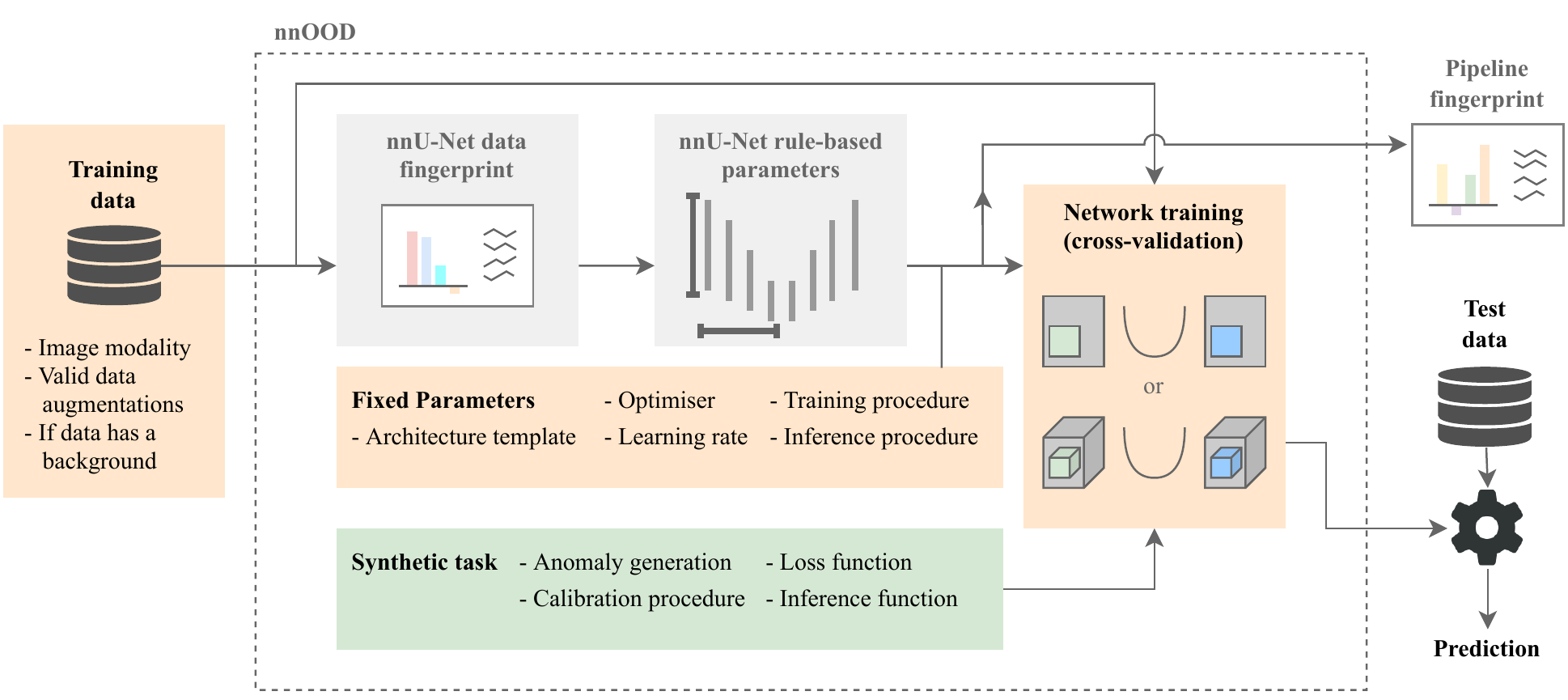}
    \caption{Overview of the nnOOD framework. The green components are entirely new to nnOOD, orange components differ significantly from their nnU-Net counterparts and grey components have only minor changes. }
    \label{fig:pipeline-overview}
\end{figure}

 In this section we discuss adaptations to the well-known nnU-Net pipeline that make it suitable for anomaly detection.
 Graphically summarized in Figure~\ref{fig:pipeline-overview}, our method differs significantly from others such as nnDetection~\cite{nnDetection} because we aim for pixel-wise predictions, allowing for a greater overlap with the original nnU-Net pipeline.

One of the primary challenges of applying the nnU-Net framework to anomaly detection is overfitting \emph{to the synthetic task}.
This is expected, as nnU-Net is intended for use on the same task at training and test time.
Therefore, it is designed to perform its training task as accurately as possible, using heavy data augmentation to avoid overfitting to the specific training data.
One way we reduce overfitting to a synthetic task is in model selection.
Originally, nnU-Net uses a five-fold cross-validation process across three different network configurations: a 2D U-Net, a 3D U-Net, and, if the data is sufficiently large, a 3D U-Net cascade.
Then the final model is chosen as the model or ensemble of two models which achieves the highest mean foreground dice score on the training set cross-validation.
In our framework we do not assume we have access to a validation set of real anomalies, and we do not want to select a model based on performance on synthetic data.
Instead, we select a model which matches the dimensionality of the data, i.e. using n-dimensional models for n-dimensional data, which performed well in nnU-Net's experiments.
We also omit the cascade setting because the dynamic nature of self-supervised tasks is less amenable to cascaded training.

Another area that requires restrictions is the training regime. The original nnU-Net training schedule is much longer than the training duration seen in many self-supervised methods.
For instance, the training schedule in Schl\"{u}ter et al.~\cite{schluter2021self} terminates well before the earliest stopping point of nnU-Net, after less than a third of the number of iterations.
To mitigate this we add early stopping based on a moving average of the average precision score (AP) on the synthetic data validation set.
We found that a threshold of 0.875 allows the models to learn useful features without overfitting to the fine-grain details of the synthetic tasks.
If a model fails to reach this threshold we utilise the original nnU-Net early stopping based on the loss plateauing as a backup.
We train each method to an equal level of validation performance (on its own task) so that we always learn each task to the same extent, regardless of it's difficulty.
By contrast, using a set number of epochs for every task would likely lead to overtraining for simple tasks, while undertraining on more complex ones.

Although data augmentation is one of the key factors for nnU-Net's success, we cannot na\"ively apply it, as many of the augmentations carry the risk of moving the sample out of the distribution of the normal data.
By learning invariance to an augmentation, the representation is trained to ignore it, preventing it from being identified as anomalous.
For example, a model trained with random rotations and translations would see misaligned data as normal.
Instead we opt to allow the user to define which augmentations are ``safe" as part of the dataset description.
This allows us to utilise as much data augmentation as possible within valid parameters for the given dataset.

The appearance of normal data can vary vastly depending on the location within the image. 
This leads to an inconsistent training signal for the model when attempting to learn to identify normality within incomplete patches.
This is particularly true when applying synthetic tasks such as CutPaste, where the patch is anomalous specifically because it appears at a different location within the image.
To allow the model to perceive the spatial context when evaluating a patch we incorporate a positional encoding, similar to ConvCoord \cite{liu2018intriguing}.
An additional channel is concatenated per spatial dimension of the data, with values ranging from -1 to 1, representing the coordinate value of that pixel for that dimension.
This has been shown to assist convergence in other anomaly detection tasks \cite{song2021anomaly}.

When selecting a patch to train on, nnU-Net chooses uniformly from all patches that lie fully within the image bounds.
This leads to the model rarely learning from regions towards the edges of the images, as only a small proportion of valid locations include them.
This is an issue at inference time where many patches are taken from the boundaries of the image, which are then more likely to be considered anomalous, due to the infrequency of observing them during training.
To rectify this, we simply randomly select from the inference patch locations at training time.
As the synthetic anomalies are often quite small relative to the size of the image, during training we oversample the anomalies.
For 30\% of each batch we choose the random patch location such that the centre of at least one anomaly was present within the patch.

When integrating the synthetic task into the framework we want to be as flexible as possible, to avoid pigeonholing future tasks into the structure of the current ones.
Formally, we define a synthetic anomaly task as:

\begin{align}
    \widetilde{x}_i, \widetilde{y}_i = f_\theta(x_i, x_j, [m_i, m_j]) && \theta = g(X, P)
    \label{equation:self-sup-task}
\end{align}

where $f$ is the anomaly generating function, producing an augmented sample $\widetilde{x}_i$ with pixel-wise label $\widetilde{y}_i$ from two samples of the distribution of normal data $x_i$ and $x_j$.
$m_i$ and $m_j$ denote the foreground masks for the corresponding samples, which are provided if present.
These are created by applying a simple Sobel operator on the image, calculating the magnitude of the gradient and using region growing from a number of locations around the image corners.
We chose to use the image gradient to determine the background as different modalities have different background intensities, but most will be uniform with a low gradient magnitude. 
Hence we only apply this if the dataset description indicates that the dataset has a consistent, uniform background.
An example of this would be in brain MRI, where the background occupies a large portion of the image, but we would not attempt to generate foreground masks for X-ray data, such as ChestX-ray14, as they are normally already cropped to the region of interest.
The parameters of the task, $\theta$, are determined by a calibration function $g$ when applied to the dataset of normal data $X$ and the current experiment plan $P$.
The calibration function is necessary because although some tasks have constant parameters, others (such as NSA \cite{schluter2021self}) use different parameters depending on the dataset.
We place no further restrictions on the implementation of these functions, allowing users to utilise or ignore the provided arguments as they see fit to produce a useful synthetic anomaly task.

To provide more flexibility, we allow the user to define the loss function $\mathcal{L}_\theta$ used when training with their task, which is given the raw network logits and the synthetic label $\widetilde{y}_i$ as input.
They can similarly define the inference function $h_\theta$, which is applied to the network logits at test time to produce pixel-wise anomaly scores.

Finally, if the synthetic task happens to follow the structure of patch-based methods such as FPI and CutPaste, we provide a compartmentalised framework to build such tasks.
This divides the task into the initial creation of the patch shape, a sequence of patch transformations, which can be spatial or alter the pixel values, the blending of the patch into the destination image and the labelling of the resulting image.
This makes it much easier to tweak existing tasks and isolate the factors that contribute the most to performance.

\section{Experiments and Results}

\noindent\textbf{Synthetic Tasks: } \label{example_tasks} As an initial baseline for future methods we have implemented the FPI \cite{tan2020detecting}, CutPaste \cite{li2021cutpaste}, PII \cite{PII} and NSA \cite{schluter2021self} tasks using our framework.
Due to their simplicity, FPI and PII can be directly generalised to arbitrary dimensions and applied to any sort of input, however the specialised nature of the other two methods requires some changes.
For CutPaste this is primarily due to their use of colour jitter, which applied brightness, contrast, saturation and hue transformations.
We omit the saturation and hue operations as the concepts do not translate to other modalities.
For contrast, we move the pixel intensity values to the mean across each channel (as opposed to the weighted average used for colour images) and for brightness we take the global minimum of the dataset as the zero value for scaling. 
Adapting NSA was a more involved task due to the number of hyperparameters that were originally chosen for each dataset by visual inspection of the generated anomalies.
At a high level, we base the maximum number of anomalies, bounds on the size of each dimension, and the minimum object area included in the extracted patch on the average foreground dimensions (treating the entire image as foreground if no background is present).
NSA also converts absolute intensity differences into labels that conform to a logistic function.
To automate this, we create 100 anomalies and calculate shape and scale parameters such that anomalous regions translate to labels with a lower bound of 0.1 that saturate at the 40th percentile.
We experimented with using both source and mixed gradients, denoted as NSA and NSA\textsubscript{Mixed} respectively.

\noindent\textbf{Data: } We evaluate these tasks on ChestX-ray14 \cite{wang2017chestxray}, a public chest X-ray dataset covering 14 common thorax disease categories as well as healthy samples.
We use the same training distribution as \cite{PII}: posteroanterior (PA) views of healthy adult patients, divided by gender to create two healthy training datasets, with 17,852 male and 14,720 female samples respectively.
For the test dataset we perform the same filtering on the unhealthy data, but further restrict it to samples that provide pathology bounding boxes.
This leaves us with 245 male and 217 female test samples.
As there are no pixel-wise annotations provided we treat the bounding boxes as anomaly masks.
For preprocessing each sample is normalised to have zero mean and unit standard deviation.
Note that we did not need to resize the images due to the patch-based nature of our framework.

\begin{figure}[ht]
    \centering
    \begin{subfigure}[t]{0.95\textwidth}
        \includegraphics[width=\linewidth]{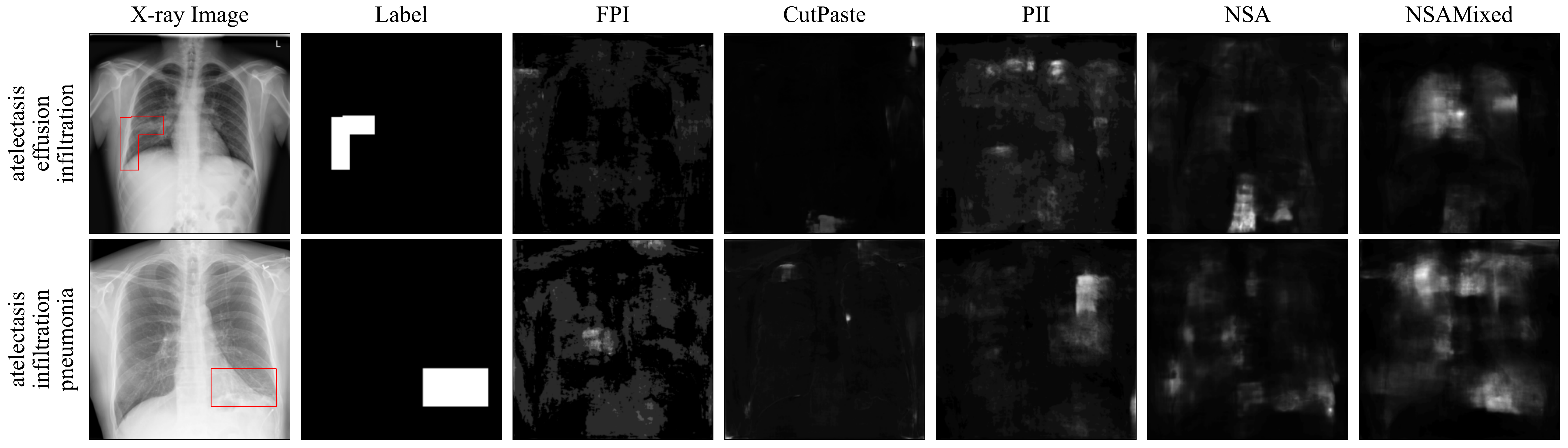}
        \caption{Male posteroanterior dataset}
    \end{subfigure}
    \begin{subfigure}[t]{0.95\textwidth}
        \includegraphics[width=\linewidth]{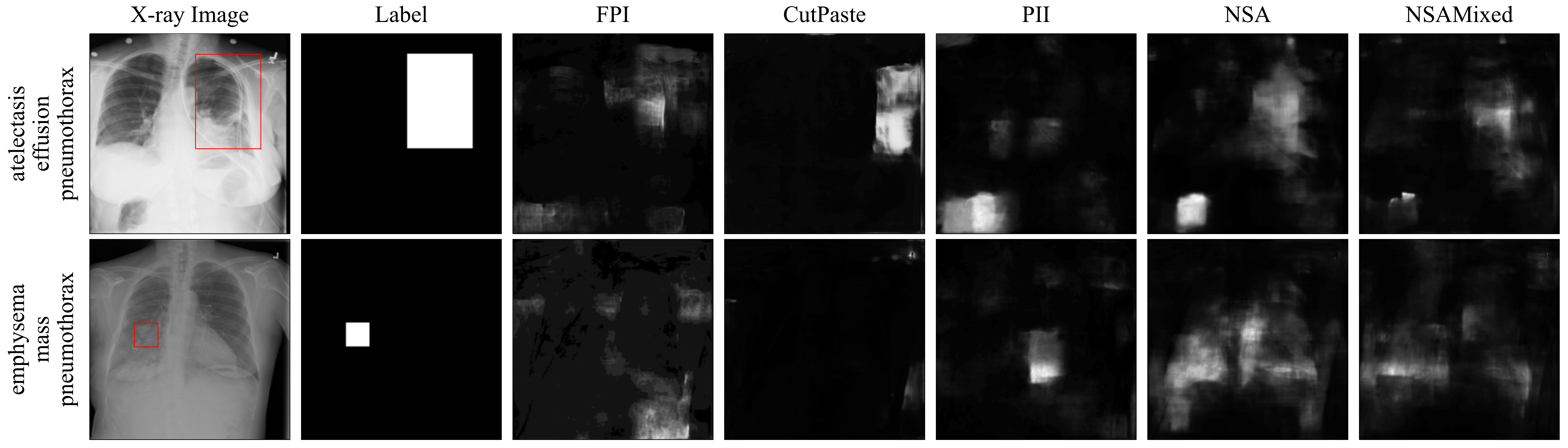}
        \caption{Female posteroanterior dataset}
    \end{subfigure}
    \caption{Examples of test predictions on each X-ray dataset. The disease labels are keywords extracted from the sample's radiologist report \cite{wang2017chestxray}.}
    \label{fig:test_visualisations}
\end{figure}

\noindent\textbf{Results: } Table~\ref{tab:main_results} shows our comparison of the different self-supervised tasks, 
with Fig.~\ref{fig:test_visualisations} displaying example test images and their predicted anomaly maps.
These scores demonstrate the challenging nature of the dataset.
Inexact ground truth bounding boxes and class imbalance make it difficult to achieve high pixel-level average precision (calculated using \texttt{scikit-learn} \cite{scikit-learn}).
For reference, a random classifier (0.5 AUROC) would achieve 0.074 and 0.063 AP for the male and female datasets respectively. 

\begin{table}[ht]
    \centering
    \caption{Pixel-wise metrics comparing models trained with different anomaly tasks using the nnOOD framework. AUROC - Area Under the Receiver Operating Characteristic curve, AP - Average Precision score}
    \begin{tabular}{lcccccccccc}
    \toprule
    Dataset & \multicolumn{5}{c}{Male PA} & \multicolumn{5}{c}{Female PA} \\
    Task    & FPI & CutPaste &    PII &    NSA & NSA\textsubscript{Mixed} & FPI & CutPaste &    PII &    NSA & NSA\textsubscript{Mixed} \\
    \cmidrule(lr){2-6}\cmidrule(lr){7-11}
    AUROC $\uparrow$   &    0.515 & 0.484 &  0.554 &  0.718 & 0.714 &  0.490 & 0.446 &  0.615 &  0.699 & 0.698 \\
    AP $\uparrow$ &         0.075 & 0.071 &  0.084 &  0.162 & 0.167 &  0.064 & 0.060 &  0.086 &  0.139 & 0.133 \\
    \bottomrule
    \end{tabular}
    \label{tab:main_results}
\end{table}

The FPI and CutPaste tasks do not seem to help the models identify medical anomalies.
This is most likely because sharp, image-aligned discontinuities are unlikely to appear in real pathologies.
Both of these methods generally predict low scores across the images (Fig.~\ref{fig:test_visualisations}), resulting in performance similar to that of a random classifier.
On the other hand, tasks which seamlessly blend their synthetic anomalies into the target image (PII, NSA, NSA\textsubscript{Mixed}) help more.
Although these approaches may not reach supervised performance, they are able to learn useful features without any exposure to real anomalies.

Interestingly, the use of different synthetic tasks massively altered the time taken to reach the AP threshold.
The difference in average training times reflects how easily the anomalies can be seen: CutPaste uses very obvious anomalies (27.3 epochs), NSA and NSA\textsubscript{Mixed} blend their patches more seamlessly (88.5 and 119.7 epochs), by not moving the extracted patch FPI's anomalies are more subtle (272.5 epochs), and PII's addition of Poisson image blending to that formula increases the subtlety even further (312.7 epochs).

\section{Conclusion}

In this paper we present a framework that makes self-supervised anomaly localisation more accessible and facilitates evaluation on a unified platform.
By automating the training configuration independently from the synthetic task, we are able to compare the true ability of each method under more controlled settings and free from unequal hyperparameter tuning.
Using this framework, we compare the current state-of-the-art methods and show that there is still much room for improvement.
We hope that nnOOD will enable further investigation of self-supervised, synthetic anomaly localisation methods across a wider variety of modalities.
Our modular design also serves as a foundation for continued research in paradigms other than patch blending, such as using deformations.

In our experiments, we focused on anomaly localisation at the pixel level.
Although sample-level detection is often reported, these scores sometimes inflate performance.
We believe that pixel-level evaluation better reflects the usefulness of these methods in clinical practice.
For example, an accurate sample-level score with poor localisation may actually mislead a clinician to pursue a tangential diagnosis.
This is particularly concerning in anomaly detection, where scores do not correspond to any specific disease classes. 
We hope that nnOOD will help facilitate future developments in anomaly detection and hold them to a higher standard, so that the field as a whole can move closer to real, beneficial tools.

\noindent\textbf{Acknowledgements:} This work was supported by the UKRI London Medical Imaging and Artificial Intelligence Centre for Value Based Healthcare.

%
%
%
\bibliographystyle{splncs04}
\bibliography{references}
\newpage

\end{document}